\documentclass{article}

\usepackage[final]{neurips_2020_arxiv}

\usepackage[utf8]{inputenc} 
\usepackage[T1]{fontenc}    
\usepackage{hyperref}       
\usepackage{url}            
\usepackage{booktabs}       
\usepackage{amsfonts}       
\usepackage{nicefrac}       
\usepackage{microtype}      
\usepackage{algorithm}
\usepackage{algorithmic}

\usepackage{microtype}
\usepackage{graphicx}
\usepackage{subfigure}
\usepackage{tikz}
\usepackage{amsmath,amssymb}
\usepackage{mathtools}
\usepackage{xfrac}

\usepackage{amsmath,amsfonts,bm}

\def\1{\bm{1}}

\DeclareMathAlphabet{\mathsfit}{\encodingdefault}{\sfdefault}{m}{sl}
\SetMathAlphabet{\mathsfit}{bold}{\encodingdefault}{\sfdefault}{bx}{n}

\newcommand{\E}{\mathbb{E}}

\newcommand{\Cov}{\mathrm{Cov}}

\DeclareMathOperator*{\argmax}{arg\,max}

\let\ab\allowbreak

\newcommand{\adagrad}{AdaGrad}
\newcommand{\rmsprop}{RMSprop}
\newcommand{\adam}{Adam}
\newcommand{\adamw}{AdamW}

\renewcommand{\L}{\mathcal{L}}

\newcommand{\muprior}{\mu_\text{prior}}
\newcommand{\mupost}{\mu_\text{post}}

\newcommand{\mprior}{\mu_\text{prior}}
\newcommand{\mpost}{\mu_\text{post}}

\newcommand{\mat}[1]{\mathbf{#1}}

\newcommand{\sprior}[1][2]{\sigma^{#1}_\text{prior}}
\newcommand{\spost}[1][2]{\sigma^{#1}_\text{post}}

\newcommand{\lpost}{\lambda_\text{post}}
\newcommand{\dd}[2][]{\frac{\partial #1}{\partial #2}}
\newcommand{\N}{\mathcal{N}\b}
\newcommand{\m}{\boldsymbol{\mu}}
\renewcommand{\P}[2][]{\operatorname{P}_{#1}\b{#2}}

\renewcommand{\sb}[1]{\left[ #1 \right]}

\renewcommand{\b}[1]{\left( #1 \right)}

\newcommand{\w}{\mat{w}}

\newcommand{\lrsgd}{\eta_{\text{SGD}}}
\newcommand{\lradam}{\eta_{\text{Adam}}}

\renewcommand{\H}{\mathbf{H}}

\newcommand{\x}{\boldsymbol{\xi}}

\renewcommand{\ab}[1]{\langle #1 \rangle}

\title{Bayesian filtering unifies adaptive and non-adaptive neural network optimization methods}

\author{%
  Laurence Aitchison\\
  Department of Computer Science\\
  University of Bristol\\
  Bristol, UK, BS8 1UB\\
  \texttt{laurence.aitchison@bristol.ac.uk}
}

\begin{document}

\maketitle

\begin{abstract}
We formulate the problem of neural network optimization as Bayesian filtering, where the observations are the backpropagated gradients.
While neural network optimization has previously been studied using natural gradient methods which are closely related to Bayesian inference, they were unable to recover standard optimizers such as Adam and RMSprop with a root-mean-square gradient normalizer, instead getting a mean-square normalizer.
To recover the root-mean-square normalizer, we find it necessary to account for the temporal dynamics of all the other parameters as they are geing optimized.
The resulting optimizer, AdaBayes, adaptively transitions between SGD-like and Adam-like behaviour, automatically recovers AdamW, a state of the art variant of Adam with decoupled weight decay, and has generalisation performance competitive with SGD.
\end{abstract}


\section{Introduction and Background}

The cannonical non-adaptive neural network optimization method is vanilla stochastic gradient descent (SGD) with momentum which updates parameters by multiplying the exponential moving average gradient, $\ab{g(t)}$, by a learning rate, $\lrsgd$,
\begin{align}
  \label{eq:def:sgd}
  \Delta w_\text{SGD}(t) &= \lrsgd \frac{\ab{g(t)}}{\text{minibatch size}}.
\end{align}
Here, we divide by the minibatch size because we define $g(t)$ to be the gradient of the \textit{summed} loss, whereas common practice is to use the gradient of the \textit{mean} loss.
Further, following the convention established by Adam \citep{kingma2014adam}, $\ab{g(t)}$, is computed by debiasing a raw exponential moving average, $m(t)$,
\begin{align}
  m(t) &= \beta_1 m(t-1) + \b{1-\beta_1} g(t) & \ab{g(t)} &= \frac{m(t)}{1-\beta_1^t}.
\end{align}
where $g(t)$ is the raw minibatch gradient, and $\beta_1$ is usually chosen to be 0.9.
These methods typically give excellent generalisation performance, and as such are used to train many state-of-the-art networks (e.g. ResNet \citep{he2016deep}, DenseNet \citep{huang2017densely}, ResNeXt \citep{xie2017aggregated}).

Adaptive methods change the learning rates as a function of past gradients.
These methods date back many years \citep[e.g.\ vario-eta][]{neuneier1998train}, and many variants have recently been developed, including \adagrad{} \citep{duchi2011adaptive}, \rmsprop{} \citep{hinton2012overview} and \adam{} \citep{kingma2014adam}.
The cannonical adaptive method, \adam{}, normalises the exponential moving average gradient by the root mean  square of past gradients,
\begin{align}
  \label{eq:def:rmsprop}
  \Delta w_\text{Adam}(t) &= \lradam \frac{\ab{g(t)}}{\sqrt{\ab{g^2(t)}}}.
\end{align}
where,
\begin{align}
  v(t) &= \beta_2 v(t-1) + \b{1-\beta_2} g^2(t)  & \ab{g^2(t)} &= \frac{v(t)}{1-\beta_2^t},
\end{align}
and where $\beta_2$ is typically chosen to be 0.999.
These methods are often observed to converge faster, and hence may be used on problems which are more difficult to optimize \citep{graves2013generating}, but can give worse generalisation performance than non-adaptive methods \citep{keskar2017improving,loshchilov2017fixing,wilson2017marginal,luo2019adaptive}.

Obtaining a principled theory of these types of method is important, as it should enable us to develop improved adaptive optimizers.
As such, here we formulated Bayesian inference as an optimization problem \citep{puskorius1991decoupled,sha1992optimal,puskorius1994neurocontrol,puskorius2001parameter,feldkamp2003simple,ollivier2017online}, and carefully considered how the dynamics of optimization of the other parameters influences any particular parameter.
We were able to recover the standard root-mean-square normalizer for RMSprop and Adam.
Critically, our approach was also able to recover a state-of-the-art variant of Adam with ``decoupled'' weight decay \citep{loshchilov2017fixing}.
As such, we hope that by pursuing our dynamical Bayesian approach further, it will be possible to develop improved adaptive optimization algorithms.



\section{Related work}

Previous work has considered the relationships between adaptive stochastic gradient descent methods and variational online Newton (VON), which is very closely related to natural gradients \citep{khan2017conjugate,khan2017vprop,khan2018fast} and Bayes \citep{ollivier2017online}.
Critically, this work found that direct application of VON/Bayes gives a sum-squared normalizer, as opposed to a root-mean-squared normalizer as in Adam and RMSProp.
In particular, see Eq.~7 in \citet{khan2018fast}, which gives the Variational-online Newton (VON) updates, and includes a mean-squared gradient normalizer.
To provide a method that matches Adam and RMSProp more closely, they go on to provide an ad-hoc modification of the VON updates, with a root-mean-square normalizer, saying ``Using ... an additional modification in the VON update, we can make the VON update very similar to RMSprop. Our modification involves taking the square-root over $\mathbf{s}(t+1)$ in Eq.~(7)''.
In contrast, our approach gives the root-mean-square normalizer directly, without any additional modifications, and automatically recovers decoupled weight decay \citep{loshchilov2017fixing} which is not recovered by VON \citep[again, see Eq.~7 in][]{khan2018fast}.

An alternative view on these results is given by considering equivalence of online natural gradients and Kalman filtering \citep{ollivier2017online}.
Through this equivalence, they have the same issues as in \citep{khan2017conjugate,khan2017vprop,khan2018fast}: having a mean-square rather than root-mean-square form for the gradient normalizer.
Further, note that while they do consider a ``fading memory'' approach, they ``multiply the log-likelihood of previous points by a forgetting factor $(1-\lambda_t)$ before each new observation. 
This is equivalent to an additional step $\mathbf{P}_{t-1} \rightarrow \mathbf{P}_{t-1}/(1 - \lambda_t)$ in the Kalman filter, or to the addition of an artificial process noise $\mathbf{Q}_t$ proportional to $\mathbf{P}_{t-1}$'', where $\mathbf{P}_{t-1}$ is their posterior covariance matrix.
Critically, their ``artifical process noise ... proportional to $\mathbf{P}_{t-1}$'' again gives a mean-square form for the gradient normalizer (see Appendix~\ref{sec:si:ollivier} for details).
In contrast, we give an alternative motivation for the introduction of \textit{fixed} process noise, and show that fixed process noise recovers the root-mean-square gradient normalizer in Adam.

\section{Methods}

\begin{figure}
  \centering
  \includegraphics[width=0.7\textwidth]{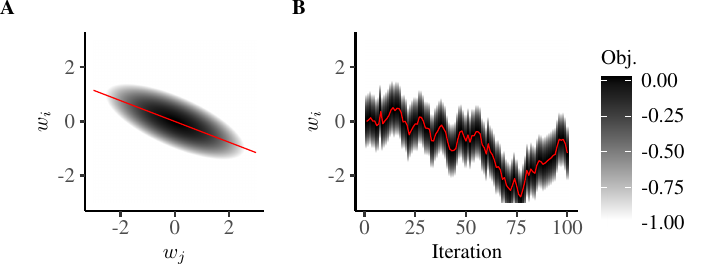}
  \caption{
    A schematic figure showing correlation-induced dynamics.
    \textbf{A} The objective function (usually equivalent to a posterior over the parameters) induces correlations between the parameter of interest, $w_i$, and other parameters (here represented by $w_j$).
    The red line displays the optimal value for $w_i$ as a function of $w_j$ or time.
    \textbf{B} The other parameters (including $w_j$) change over time as they are also being optimized, implying that the optimal value for $w_i$ changes over iterations.
    \label{fig:schematic}
  }
\end{figure}

Here, we set up the problem of neural network optimization as Bayesian inference.
Typically, when performing Bayesian inference, we would like to reason about correlations in the full posterior over all parameters jointly (Fig.~\ref{fig:schematic}A).
However, neural networks have so many parameters that reasoning about correlations is intractable: instead, we are forced to work with factorised approximate posteriors.
To understand the effects of factorised approximate posteriors, consider the $i$th parameter.
The current estimate of the other parameters, $\m_{-i}(t)$ changes over time, $t$, as they are optimized. 
As there are correlations in the posterior, the optimal value for the $i$th parameter, $w_i^*(t)$, conditioned on the current setting of the other parameters, $\m_{-i}(t)$ also changes over time (Fig.~\ref{fig:schematic}B),
\begin{align}
  w_i^*(t) = \argmax_{w_i} \mathcal{L}\b{w_i, \m_{-i}(t)}.
\end{align}
As such, to form optimal estimates, $\mu_i(t)$ of the $i$th parameter, we need to reason changes over time in the optimal setting for that parameter, $w_i^*(t)$.
If we knew the full, correlated posterior in Fig.~\ref{fig:schematic}A, then we could compute the change in the $i$th parameter from the change in all the other parameters.
However, in our case, the correlations are unknown, so the best we can do is to say that the new optimal value for the $i$th parameter will be close to --- but slightly different from --- the current optimal value, and these changes in the optimal value (Fig.~\ref{fig:schematic}B) constitute stochastic-dynamics that are implicitly induced by our choice of a factorised approximate posterior.

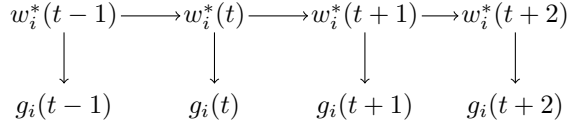
\begin{figure}
  \centering
  \begin{tikzpicture}
    \def\dx{2cm}
    \def\dy{1.2cm}

    \node[inner sep=1pt] (wm1) at ({-1*\dx}, \dy) {$w^*_i(t-1)$};
    \node[inner sep=1pt] (w)   at ({     0}, \dy) {$w^*_i(t)$};
    \node[inner sep=1pt] (wp1) at ({ 1*\dx}, \dy) {$w^*_i(t+1)$};
    \node[inner sep=1pt] (wp2) at ({ 2*\dx}, \dy) {$w^*_i(t+2)$};

    \node[] (gm1) at ({-1*\dx}, 0) {$g_i(t-1)$};
    \node[] (g)   at ({     0}, 0) {$g_i(t)$};
    \node[] (gp1) at ({ 1*\dx}, 0) {$g_i(t+1)$};
    \node[] (gp2) at ({ 2*\dx}, 0) {$g_i(t+2)$};

    \draw[->] (wm1) -- (w);
    \draw[->] (w)   -- (wp1);
    \draw[->] (wp1) -- (wp2);

    \draw[->] (wm1) -- (gm1);
    \draw[->] (w)   -- (g);
    \draw[->] (wp1) -- (gp1);
    \draw[->] (wp2) -- (gp2);
  \end{tikzpicture}
  \caption{
    Graphical model under which we perform inference.
    \label{fig:graphical-model}
  }
\end{figure}

More formally, we can explicitly consider the stochastic dynamics of $w_i^*(t)$ that emerge under a quadratic objective.
We take the objective for a single datapoint (which, for a supervised learning problem, would be a single input, $\mathbf{x}_\alpha$, output, $y_\alpha$, pair) to be quadratic,
\begin{align}
  \label{eq:def:L}
  \L\b{\mathbf{x}_\alpha, y_\alpha; \w} = \L_\alpha\b{\w} &= - \tfrac{1}{2} \w^T \H \w + \x_\alpha^T \w,
\end{align}
where we have assumed that the Hessian is the same across data points, but the location of the mode varies across datapoints.
Using the Fisher Information identity (see Appendix~\ref{sec:si:fi}), we can identify the covariance of the datapoint-dependent noise term as equal to the Hessian,
\begin{align}
  \label{eq:cov:xi}
  \E\sb{\x_\alpha} &= \mathbf{0} & \Cov\sb{\x_\alpha} &= \H.
\end{align}
The gradient for a single datapoint is 
\begin{align}
  \dd{\w} \L_\alpha\b{\w}  &= - \H \w + \x_\alpha 
\end{align}
Thus, the gradient with respect to the $i$th weight, when all the other parameters, $\w_{-i}$ are set to the current estimate, $\m_{-i}(t)$, is
\begin{align}
  \dd{w_i}\L_\alpha\b{w_i, \m_{-i}(t)} = -H_{ii} w_i - \H_{-i, i}^T \m_{-i}(t) + \xi_{\alpha,i}
\end{align}
where $\H_{-i, i}^T$ is the $i$th column of the Hessian, omitting the $ii$th element.
The optimal value for the $i$th parameter can be found by solving for the value of $w_i$ for which the gradient of the average objective is zero, 
\begin{align}
  \label{eq:def:w*}
  w_i^*(t) &= -\frac{1}{H_{ii}} \H_{-i, i}^T \m_{-i}(t).
  \intertext{Thus, we can rewrite the gradients as a function only of the optimal weight,}
  \dd{w_i}\L_\alpha\b{w_i, \m_{-i}(t)} &= H_{ii} \b{w_i^*(t) - w_i} + \xi_{\alpha, i}.
\end{align}
The data we actually measure is the gradient of the objective evaluated at the current estimate of the parameter, $w_i = \mu_i$. Taking into account the stochasticity across datapoints given by $\xi_{\alpha,i}$, the gradient has distribution,
\begin{align}
  \label{eq:def:like}
  \P{g_i(t)| w_i^*(t)} &= \N{H_{ii} \b{w_i^*(t) - \mu_i(t)}, H_{ii}}.
\end{align}
This expression is highly suggestive: we should be able to infer the optimal value for the $i$th parameter, $w_i^*(t)$, from the backpropagated gradients, $g_i(t)$.
However, to do this inference correctly, we need to understand the stochastic dynamics of $w_i^*(t)$.
In particular, Eq.~\eqref{eq:def:w*} shows us that the dynamics of $w_i^*(t)$ are governed by the dynamics of our estimates, $\m_{-i}(t)$, as they are optimized, and this optimization is a complex stochastic process.
As reasoning about the dynamics of $\m_{-i}(t)$ under optimization is intractable, we instead consider simpler, discretised Ornstein-Uhlenbeck dynamics for $\m_{-i}(t)$,
\begin{align}
  \P{\m_{-i}(t+1)| \m_{-i}(t)} &= \N{\b{1 - \tfrac{\eta^2}{2 \sigma^2}} \m_{-i}(t), \eta_\mu^2}.
\end{align}
Thus, the dynamics for the $i$th optimal weight become,
\begin{align}
  \label{eq:def:dyn}
  \P{w_i^*(t+1)| w_i^*(t)} &= \N{\b{1 - \tfrac{\eta^2}{2 \sigma^2}} w_i^*(t), \eta^2}
  \intertext{where,}
  \eta^2 &= \eta_\mu^2 \; \H_{-i, i}^T \H_{-i, i}.
\end{align}
Combined, Eq.~\eqref{eq:def:like} and Eq.~\eqref{eq:def:dyn} define a stochastic linear dynamical system where the optimal weight, $\w_i^*(t)$ is the latent variable, and the backpropagated gradients, $g_i(t)$ are the observations (Fig.~\ref{fig:graphical-model}).

As such, we are able to use standard Kalman filter updates to infer a distribution over $w^*(t)=w_i^*(t)$ (where we drop indices for brevity).
As the dynamics and likelihood are Gaussian, the Kalman filter priors and posteriors are,
\begin{subequations}
\begin{align}
  \label{eq:def:prior}
  \P{w^*(t)| g(t-1),\dotsc,g(1)} &= \N{\mprior(t), \sprior(t)},\\
  \label{eq:def:post}
  \P{w^*(t)| g(t),\dotsc,g(1)} &= \N{\mpost(t), \spost(t)},
\end{align}
\end{subequations}
where we evaluate the gradient, $g(t)=g_i(t)$ at $\mu_i(t) = \mprior(t)$, and where the updates for $\mprior(t)$ and $\sprior(t)$ can be computed from Eq.~\eqref{eq:def:dyn},
\begin{subequations}
  \label{eq:prior}
  \begin{align}
    \label{eq:def:mprior}
    \muprior(t) &= \b{1-\tfrac{\eta^2}{2 \sigma^2}}\phantom{^2} \mupost(t-1),\\
    \label{eq:def:Sprior}
    \sprior(t) &= \b{1-\tfrac{\eta^2}{2 \sigma^2}}^2 \spost(t-1) + \eta^2.
  \end{align}
\end{subequations}
And the updates for $\mpost(t)$ and $\spost(t)$ come from applying Bayes theorem (Appendix~\ref{sec:si:kf}), with the likelihood given by Eq.~\eqref{eq:def:like}. 
Following the standard approach in this line of work \citep{khan2017conjugate,zhang2017noisy,khan2017vprop,khan2018fast}, we approximate $H_{ii}$ using the squared gradient (improving upon this approximation is an important avenue for future work, but not our focus here),
\begin{subequations}
  \label{eq:post}
  \begin{align}
    \label{eq:def:Spost}
    \spost(t) &= \frac{1}{\frac{1}{\sprior(t)} + H_{ii}} \approx \frac{1}{\frac{1}{\sprior(t)} + g^2(t)},\\
    \label{eq:def:mpost}
    \mupost(t) &= \muprior(t) + \spost(t) g(t).
  \end{align}
\end{subequations}
The full updates are now specified by iteratively applying Eq.~\eqref{eq:prior} and Eq.~\eqref{eq:post}.

\begin{figure*}
  \centering
  \includegraphics{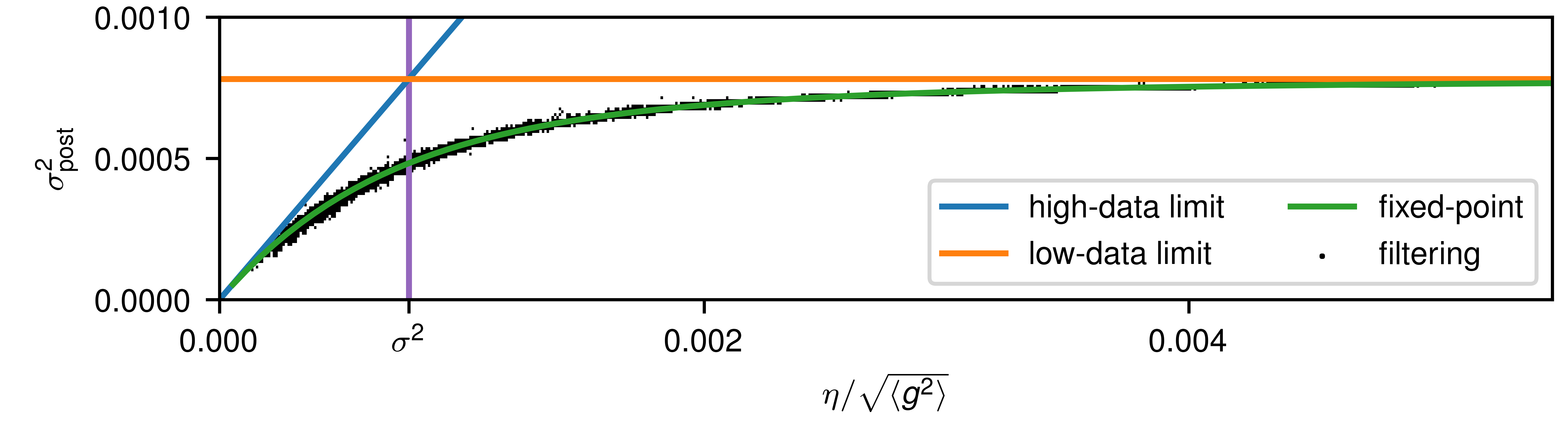}
  \caption{
    The learning rate for AdaBayes (points) compared against the predicted fixed-point value (green line), $\spost$.
    The plot displays the low-data limit (orange line), which is valid when the value on the x-axis, $\eta / \sqrt{\ab{g^2}}$, is much greater than $\sigma^2$ (purple line), and the high-data limit (blue line), which is valid when the value on the x-axis is much smaller than $\sigma^2$ (purple line).
    \label{fig:approx}
  }
\end{figure*}

Next, we make two minor modifications to the updates for the mean, to match current best practice for optimizing neural networks.
First, we allow more flexibility in weight decay, by replacing the $\eta^2/(2\sigma^2)$ term in Eq.~\eqref{eq:def:mprior} with a new parameter, $\lambda$.
Second, we incorporate momentum, by using an exponential moving average gradient, $\ab{g(t)}$, instead of the raw minibatch gradient in Eq.~\eqref{eq:def:mpost}.
In combination, the updates for the mean become,
\begin{subequations}
  \label{eq:def:flex} 
  \begin{align}
    \label{eq:def:mprior:flex}
    \muprior(t) &= \b{1-\lambda}\phantom{^2} \mupost(t-1),\\
    \label{eq:def:mpost:flex}
    \mupost(t) &= \muprior(t) + \spost(t) \ab{g(t)}.
  \end{align}
\end{subequations}
Our complete Bayesian updates are now given by using Eq.~\eqref{eq:def:flex} to update $\muprior$ and $\mupost$, and using Eq.~\eqref{eq:def:Sprior} and Eq.~\eqref{eq:def:Spost} to update $\sprior$ and $\spost$ (see Algo.~\ref{algo:adabayes}).


\subsection{AdaBayes recovers SGD and \adam{}}
To understand how AdaBayes relates to previous algorihtms (SGD and \adam{}), we plotted the AdaBayes learning rate, $\spost$ against the \adam{} learning rate, $\eta/\sqrt{\ab{g^2}}$ (Fig.~\ref{fig:approx}, points) for the ResNet-34 considered later.
We found that for high values of $\ab{g^2}$, corresponding to large values of the Fisher Information, the AdaBayes learning rate closely matched the \adam{} learning rate (Fig.~\ref{fig:approx}, blue line).
In contrast, as the value of $\ab{g^2}$ decreased, corresponding to smaller values of the Fisher Information, we found that the AdaBayes learning rate became constant, mirroring standard SGD (Fig.~\ref{fig:approx}, orange line).
Thus, Fig.~\ref{fig:approx} empirically establishes that AdaBayes converges to SGD in the low data (Fisher-Information) limit, and \adam{} in the high data limit. 
Furthermore, the tight vertical spread of points in Fig.~\ref{fig:approx} indicates that, in practice, the AdaBayes value of $\spost$ is largely determined by the Fisher-Information, $\ab{g^2}$, thus raising the question of whether we can obtain better understanding of the relationship between $\ab{g^2}$ and $\spost$.
Indeed, such an understanding is possible, if we consider the fixed point of the $\spost$ updates (Eq.~\ref{eq:def:Sprior} and \ref{eq:def:Spost}).
%
To obtain the fixed-point, we substitute the update for $\spost$ (Eq.~\ref{eq:def:Spost}) into the update for $\sprior$ (Eq.~\ref{eq:def:Sprior}), and neglect small terms (see Appendix~\ref{sec:si:approx-cov}), which tells us that the fixed-point $\spost$ is given by the solution of a quadratic equation,
\begin{align}
  \label{eq:quad}
  0 &\approx \sigma^2 \b{\frac{1}{\spost}}^2 - \frac{1}{\spost} - \frac{\ab{g^2} \sigma^2}{\eta^2}. 
\end{align}
Solving for $1/\spost$, we obtain,
\begin{align}
  \label{eq:rms:spost}
  \frac{1}{\spost} &\approx \frac{1}{2\sigma^2} \b{1 + \sqrt{1 + 4 \b{\frac{\sigma^2}{{\nicefrac{\eta}{\mkern-5mu _{\sqrt{\langle g^2 \rangle}}}}}}^2}}.
\end{align}
We confirmed the fixed-point indeed matches the empirically measured AdaBayes learning rates by plotting the fixed-point predictions in Fig.~\ref{fig:approx} (green line).
Importantly, the fixed-point expression merely helps understand a result that we established empirically.
Finally, this close match means that we can define another set of updates, AdaBayes-FP, where we set $\spost$ directly to the fixed-point value, using Eq.~\eqref{eq:rms:spost}, rather than using the full AdaBayes updates given by Eq.~\eqref{eq:def:Sprior} and Eq.~\eqref{eq:def:Spost}.

\begin{figure*}[t]
  \begin{minipage}{0.45\textwidth}
    \begin{algorithm}[H]
      \caption{AdaBayes}
      \label{algo:adabayes}
      \begin{algorithmic}
        \STATE $\mathrlap{\eta} \phantom{\sprior} \gets \lradam$
        \STATE $\mathrlap{\sigma^2} \phantom{\sprior} \gets \lrsgd / \text{minibatch size}$
        \STATE $\sprior \gets \sigma^2$
        \WHILE{not converged}
          \STATE $\mathrlap{g} \phantom{\ab{g^2}} \gets \nabla \L_t(\mu)$ 
          \STATE $\mathrlap{m} \phantom{\ab{g^2}} \gets \beta_1 m + \b{1-\beta_1} g$
          \STATE \textcolor{white}{$\mathrlap{v} \phantom{\ab{g^2}} \gets \beta_2 \mathrlap{v} \phantom{m} + \b{1-\beta_2} g^2$}
          \STATE $\mathrlap{\ab{g}} \phantom{\ab{g^2}}  \gets m/\b{1-\beta_1^t}$
          \STATE \textcolor{white}{$\ab{g^2} \gets \mathrlap{v} \phantom{m}/\b{1-\beta_2^t}$}
          \STATE $\mathrlap{\sprior} \phantom{\ab{g^2}} \gets \b{1 - \tfrac{\eta^2}{2 \sigma^2}}^2 \spost + \eta^2$
          \STATE $\mathrlap{\spost} \phantom{\ab{g^2}} \gets \frac{1}{\sprior[-2] + g^2}$
          \STATE $\mathrlap{\mu} \phantom{\ab{g^2}} \gets \b{1-\lambda} \mu + \spost \ab{g}$
        \ENDWHILE
      \end{algorithmic}
    \end{algorithm}
  \end{minipage}
  \hfill
  \begin{minipage}{0.45\textwidth}
    \begin{algorithm}[H]
      \caption{AdaBayes-FP}
      \label{algo:adabayes-ss}
      \begin{algorithmic}[1]
        \STATE $\mathrlap{\eta} \phantom{\sprior} \gets \lradam$
        \STATE $\mathrlap{\sigma^2} \phantom{\sprior} \gets \lrsgd / \text{minibatch size}$
        \textcolor{white}{\STATE $\sprior \gets \sigma^2$}
        \WHILE{not converged}
        \STATE $\mathrlap{g} \phantom{\ab{g^2}} \gets \nabla \L_t(\mu)$ 
        \STATE $\mathrlap{m} \phantom{\ab{g^2}} \gets \beta_1 m + \b{1-\beta_1} g$
        \STATE $\mathrlap{v} \phantom{\ab{g^2}} \gets \beta_2 \mathrlap{v} \phantom{m} + \b{1-\beta_2} g^2$
        \STATE $\mathrlap{\ab{g}} \phantom{\ab{g^2}}  \gets m/\b{1-\beta_1^t}$
        \STATE $\ab{g^2} \gets \mathrlap{v} \phantom{m}/\b{1-\beta_2^t}$
        \STATE \textcolor{white}{$\mathrlap{\sigma^2} \phantom{\ab{g^2}} \gets \frac{1}{\sigma^{-2} + g^2}$}
        \STATE $\mathrlap{\spost} \phantom{\ab{g^2}} \gets \frac{1}{2\sigma^2} + \sqrt{\frac{1}{4 \sigma^4} + \frac{\ab{g^2}}{\eta^2}}$
        \STATE $\mathrlap{\mu} \phantom{\ab{g^2}} \gets \b{1-\lambda} \mu + \spost \ab{g}$
        \ENDWHILE
      \end{algorithmic}
    \end{algorithm}
  \end{minipage}
\end{figure*}

\subsubsection{Recovering SGD in the low-data limit}

In the low-data regime where $\eta / \sqrt{\ab{g^2}} \gg \sigma^2$, the empirically measured AdaBayes learning rate, $\spost$, becomes constant (Fig.~\ref{fig:approx}; orange line), so the AdaBayes updates (Eq.~\ref{eq:def:mpost:flex}) become approximately equivalent to vanilla SGD (Eq.~\ref{eq:def:sgd}).
To understand this convergence, we can leverage the fixed-point expression in Eq.~\eqref{eq:rms:spost} which accurately models empirically measured learning rates,
\begin{align}
  \label{eq:low-data}
  \lim_{\ab{g^2} \rightarrow 0} \spost &\approx \sigma^2,
\end{align}
We can leverage this equivalence to set $\sigma^2$ using standard values of the SGD learning rate,
\begin{align}
  \label{eq:sigma2}
  \sigma^2 = \frac{\lrsgd}{\text{minibatch size}}.
\end{align}
Setting $\sigma^2$ in this way would suggest $\sigma^2 \sim 0.001$\footnote{here we use $x \sim y$ as in Physics to denote ``$x$ has the same order of magnitude as $y$'', see Acklam and Weisstein ``Tilde'' MathWorld. http://mathworld.wolfram.com/Tilde.html}, as $\lrsgd \sim 0.1$, and the $\text{minibatch size} \sim 100$.
It is important to sanity check that this value of $\sigma^2$ corresponds to Bayesian filtering in a sensible generative model.
In particular, note that $\sigma^2$ is the variance of the prior over $w_i$, and as such $\sigma^2$ should correspond to typical initialization schemes \citep[e.g.][]{he2015delving} which ensure that input and output activations have roughly the same scale. 
These schemes use $\sigma^2 \sim 1/(\text{number of inputs})$, and if we consider that there are typically $\sim\!\!100$ input channels, and we typically convolve over a $3\times 3=9$ pixel patch, we obtain $\sigma^2 \sim 0.001$, matching the value we use.

\subsubsection{Recovering Adam(W) in the high-data limit}

In the high-data regime where $\eta / \sqrt{\ab{g^2}} \ll \sigma^2$, the empirically measured AdaBayes learning rate, $\spost$, approaches the \adam{} learning rate (Fig.~\ref{fig:approx}; blue line), so AdaBayes becomes approximately equivalent to Adam(W).
To understand this convergence, we can leverage the fixed-point expression in Eq.~\eqref{eq:rms:spost} which accurately models empirically measured learning rates,
\begin{align}
  \label{eq:high-data}
  \lim_{\ab{g^2} \rightarrow \infty} \spost &\approx \frac{\eta}{\sqrt{\ab{g^2}}} 
\end{align}
so the updates (Eq.\ref{eq:def:mpost:flex}) become equivalent to \adam{} updates if we take,
\begin{align}
  \eta = \lradam.
\end{align}
As such, we are able to use past experience with good values for the Adam learning rate $\lradam$, to set $\eta$: in our case we use $\eta=0.001$.

Furthermore, when we consider the form of regularisation implied by our updates, we recover a state-of-the-art variant of \adam{}, known as \adamw{} \citep{loshchilov2017fixing}.
In standard \adam{}, weight-decay regularization is implemented by incorporating an L2 penalty on the weights in the loss function, so the gradient of the loss and regularizer are both normalized by the root-mean-square gradient.
In contrast, \adamw{} ``decouples'' weight decay from the loss, such that the gradient of the loss is normalized by the root-mean-square gradients, but the weight decay is not.
To see that our updates correspond to \adamw{}, we combine Eq.~\eqref{eq:def:mprior:flex} and Eq.~\eqref{eq:def:mpost:flex}, and substitute for $\spost$ (Eq.~\ref{eq:high-data}),
\begin{align}
  \mpost(t) &\approx \b{\lambda-1} \mpost(t-1) + \frac{\eta}{\sqrt{\ab{g^2(t)}}} \ab{g(t)}.
\end{align}
Indeed, the root-mean-square normalization applies only to the gradient of the loss, as in \adamw{}, and not to the weight decay term, as in standard \adam{}.

Finally, note that AdaBayes-FP becomes \textit{exactly} AdamW when we set $\sigma^2 \rightarrow \infty$,
\begin{align}
  \lim_{\sigma^2 \rightarrow \infty} \frac{1}{\spost} &= \lim_{\sigma^2 \rightarrow \infty} \b{\frac{1}{2\sigma^2} + \sqrt{\frac{1}{4 \sigma^4} + \frac{\ab{g^2}}{\eta^2}}} = \frac{\eta}{\sqrt{\ab{g^2}}},
\end{align}
because we use the standard Adam(W) approach to computing unbiased estimates of $\ab{g}$ and $\ab{g^2}$ (see Algo.~\ref{algo:adabayes-ss}).

\section{Experiments}
\newcommand{\sgd}{\textcolor{gray}}
\newcommand{\bestissgd}[1]{\textbf{\textcolor{gray}{#1}}}
\newcommand{\bestexclsgd}{\textbf}
\newcommand{\bestisnonsgd}{\textbf}
\begin{table*}
  \caption{
    A table displaying the minimal test error and test loss for a ResNet and DenseNet applied to CIFAR-10 and CIFAR-100 for different optimizers.
    The table displays the best adaptive algorithm (bold), which is always one of our methods: either AdaBayes or AdaBayes-FP.
    We also display the instances where SGD (gray) beats all adaptive methods (in which case we also embolden the SGD value).
    \label{table}
  }
  \centering
  \begin{tabular}{lllllllll}
    \toprule
    & \multicolumn{4}{c}{CIFAR-10} & \multicolumn{4}{c}{CIFAR-100}\\
    \cmidrule(r){2-5} \cmidrule(r){6-9}
    & \multicolumn{2}{c}{ResNet} &  \multicolumn{2}{c}{DenseNet} & \multicolumn{2}{c}{ResNet} & \multicolumn{2}{c}{DenseNet}\\
    \cmidrule(r){2-3} \cmidrule(r){4-5} \cmidrule(r){6-7} \cmidrule(r){8-9}
    optimizer & error (\%) & loss & error (\%) & loss & error (\%) & loss & error (\%) & loss \\
    \midrule
    \sgd{SGD} & \sgd{5.170} & \bestissgd{0.174} & \sgd{5.580} & \sgd{0.177} & \bestissgd{22.710} & \bestissgd{0.833} & \bestissgd{21.290} & \bestissgd{0.774} \\
Adam & 7.110 & 0.239 & 6.690 & 0.230 & 27.590 & 1.049 & 26.640 & 1.074 \\
AdaGrad & 6.840 & 0.307 & 7.490 & 0.338 & 30.350 & 1.347 & 30.110 & 1.319 \\
AMSGrad & 6.720 & 0.239 & 6.170 & 0.234 & 27.430 & 1.033 & 25.850 & 1.103 \\
AdaBound & 5.140 & 0.220 & 4.850 & 0.210 & 23.060 & 1.004 & 22.210 & 1.050 \\
AMSBound & 4.940 & 0.210 & 4.960 & 0.219 & 23.000 & 1.003 & 22.360 & 1.017 \\
AdamW & 5.080 & 0.239 & 5.190 & 0.214 & 24.850 & 1.142 & 23.480 & 1.043 \\
AdaBayes-SS & 5.230 & \bestexclsgd{0.187} & 4.910 & \bestisnonsgd{0.176} & 23.120 & \bestexclsgd{0.935} & 22.600 & \bestexclsgd{0.934} \\
AdaBayes & \bestisnonsgd{4.840} & 0.229 & \bestisnonsgd{4.560} & 0.222 & \bestexclsgd{22.920} & 0.969 & \bestexclsgd{22.090} & 1.079\\
    \bottomrule
  \end{tabular}
\end{table*}

\begin{figure*}
  \includegraphics[width=0.8\textwidth]{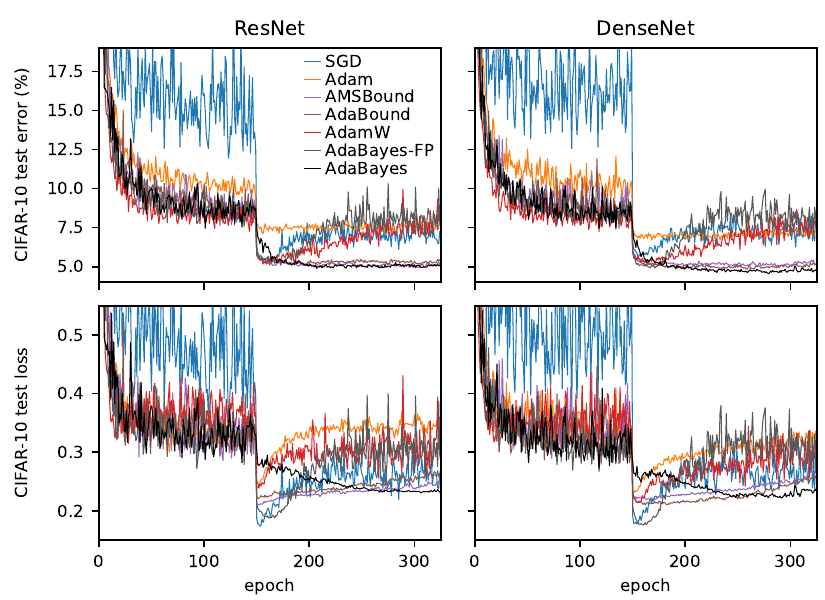}
  \centering
  \caption{Test loss and classification error for CIFAR-10 and CIFAR-100 for a Resnet-34 and a DenseNet-121, for multiple update algorithms.
    \label{fig:all}
  }
\end{figure*}

For our experiments, we have adapted the code and protocols from a recent paper \citep{luo2019adaptive} on alternative methods for combining non-adaptive and adaptive behaviour (AdaBound and AMSBound).
They considered a 34-layer ResNet \citep{he2016deep} and a 121-layer DenseNet on CIFAR-10 \citep{huang2017densely}, trained for 200 epochs with learning rates that decreased by a factor of $10$ at epoch 150.
We used the exact same networks and protocol, except that we run for more epochs, we plot both classification error and the loss, and we use both CIFAR-10 and CIFAR-100.
We used their optimized hyperparameter settings for standard baselines (including SGD and \adam{}), and their choice of hyperparameters for their methods (AdaBound and AMSBound).
For AdamW and AdaBayes, we used $\lrsgd=0.1$ and set $\sigma^2$ using Eq.~\eqref{eq:sigma2}, and we used $\lradam=\eta=0.001$ (matched to the optimal learning rate for standard \adam{}). 
We used decoupled weight decay of $5\times 10^{-4}$ \citep[from][]{luo2019adaptive}, and we used the equivalence of SGD with weight decay and SGD with decoupled weight decay to set the decoupled weight decay coefficient to $\lambda=5\times 10^{-5}$ for \adamw{}, AdaBayes and AdaBayes-FP.

The results are given in Table~\ref{table} and Fig.~\ref{fig:all}.
The best adaptive method is always one of our methods (AdaBayes or AdaBayes-FP), though SGD is frequently superior to all adaptive methods tested.
To begin, we compare our methods (AdaBayes and AdaBayes-FP) to the canonical non-adaptive (SGD) and adaptive (Adam) method (see Fig.~\ref{fig:baselines} for a cleaner figure, including other baselines).
Note that AdaBayes and AdaBayes-FP improve their accuracy and loss more rapidly than baseline methods (i.e.\ SGD and Adam) during the initial part of learning.
Our algorithms give better test error and loss than \adam{}, for all networks and datasets, they give better test error than SGD for CIFAR-10, and perform similarly to SGD in the other cases, with AdaBayes-FP often giving better performance than AdaBayes.
Next, we see that AdaBayes-FP improves considerably over AdaBayes (see Fig.~\ref{fig:adamw} for a cleaner figure), except in the case of CIFAR-10 classification error, where the difference is minimal.

Given the difficulties inherent in these types of comparison, we feel that only two conclusions can reasonably be drawn from these experiments.
First, AdaBayes and AdaBayes-SS have comparable performance to other state-of-the-art adaptive methods, including AdamW, AdaBound and AMSBound.
Second, and as expected, SGD frequently performs better than all adaptive methods, and the difference is especially dramatic if we focus on the test-loss for CIFAR-100.

\section{Conclusions}

Our fundamental contribution is show that, if we seek to use Bayesian inference to perform stochastic optimization, we need a model describing the dynamics of all the other parameters as they are optimized.
We found that even by assuming that the other parameters obey oversimplified OU dynamics, we recovered state-of-the-art adaptive optimizers (AdamW).
In our experiments, either AdaBayes or AdaBayes-FP outperformed other adaptive methods, including AdamW \citep{loshchilov2017fixing}, and Ada/AMSBound \citep{luo2019adaptive}, though SGD frequently outperformed all adaptive methods.
We hope that understanding optimization as inference, taking into account the dynamics in the other weights as they are optimized, will allow for the development of improved optimizers, for instance by exploiting Kronecker factorisation \citep{martens2015optimizing,grosse2016kronecker,zhang2017noisy}.

\bibliography{refs}
\bibliographystyle{icml2020}

\clearpage

\appendix
\renewcommand\thefigure{A\arabic{figure}}    
\setcounter{figure}{0}

\section{Mean square normalizer in Ollivier (2017)}
\label{sec:si:ollivier}
In our framework, we can encode the multiplication by the forgetting factor in the computation of $\sprior(t+1)$ from $\spost(t)$,
\begin{align}
  \label{eq:von}
  \frac{1}{\sprior(t+1)} &= \frac{1-\lambda}{\spost(t)},
\end{align}
and the equivalent process noise is,
\begin{align}
  \eta &= \frac{\lambda}{1-\lambda} \spost(t).
\end{align}
To understand the typical learning rates in this model we perform a fixed-point analysis by substituting Eq.~\eqref{eq:von} into Eq.~\eqref{eq:def:Spost},
\begin{align}
  \frac{1}{\spost(t+1)} &= \frac{1-\lambda}{\spost(t)} + g^2(t).
\end{align}
Solving for the fixed point, $\spost = \spost(t) = \spost(t+1)$, this choice of process noise gives a mean-squre normalizer,
\begin{align}
  \spost &= \frac{\lambda}{\ab{g^2}}.
\end{align}

\section{Fisher Information}
\label{sec:si:fi}
We assume that the objective for a single datapoint is a quadratic log-likelihood with a form given by Eq.~\eqref{eq:def:L},
\begin{align}
  \L\b{\mathbf{x}_\alpha, y_\alpha; \w} = \log \P{y_\alpha| \mathbf{x}_\alpha, \w}.
\end{align}
As such, if we evaluate the gradients of the objective at the correct value of $\w$ (without loss of generality, this is $\mathbf{w} = \mathbf{0}$ in Eq.~\eqref{eq:def:L}), we have,
\begin{align}
  \mathbf{g}' &= \x_\alpha.
\end{align}
And the Fisher Information identity tells us that at this location,
\begin{align}
  \label{eq:fi}
  \E\sb{\x_\alpha \x_\alpha^T} &= \E\sb{\mathbf{g}' \mathbf{g}'^T} = \H,
\end{align}
substituting the value for $\mathbf{g}'$ into this expression, we find that covariance of $\x_\alpha$ is equal to the Hessian, recovering Eq.~\eqref{eq:cov:xi}.

\section{Kalman filter}
\label{sec:si:kf}
The log-posterior (Eq.~\ref{eq:def:post}) is the sum of the log-prior (Eq.~\ref{eq:def:prior}) and the log-likelihood (Eq.~\ref{eq:def:like}).
As such,
\begin{multline}
  -\tfrac{1}{2 \spost} \b{w_i^* - \mpost}^2 = -\tfrac{1}{2 \sprior} \b{w_i^* - \mprior}^2 \\- \tfrac{1}{2 H_{ii}}\b{g_i - H_{ii}\b{w_i^* - \mprior}}^2.
\end{multline}
The quadratic terms allow us to identify $\spost$,
\begin{align}
  -\tfrac{1}{2 \spost} w_i^{*2} &= -\tfrac{1}{2 \sprior} w_i^{*2} - \tfrac{1}{2} H_{ii} w_i^{*2},\\
  \intertext{so,}
  \frac{1}{\spost} &= \frac{1}{\sprior} + H_{ii}.
  \intertext{or,}
  \spost &= \frac{1}{\tfrac{1}{\sprior} + H_{ii}}.
  \intertext{And the linear terms allow us to identify $\mpost$,}
  \tfrac{\mpost}{\spost} w_i^* &= \tfrac{\mprior}{\sprior} w_i^* + g_i w_i^* + H_{ii} \mprior w_i^*
  \intertext{so,}
  \mpost &= \spost \b{\b{\tfrac{1}{\sprior} + H_{ii}} \mprior + g_i}
  \intertext{identifing $1/\spost$,}
  \mpost &= \mprior + \spost g_i.
\end{align}

Finally, as $H_{ii}$ is unknown, we use,
\begin{align}
  H_{ii} \approx g_i^2(t),
\end{align}
as $g^2(t)$ is an unbiased estimator of $H_{ii}$ when $\mu_i(t)=\mprior(t)$ is at the optimum (Eq.~\ref{eq:fi}).

\section{Fixed point variance}
\label{sec:si:approx-cov}
For the fixed-point covariance, it is slightly more convenient to work with the inverse variance,
\begin{align}
  \lpost &= \frac{1}{\spost},
\end{align}
though the same results can be obtained through either route.
Substituting Eq.~\eqref{eq:def:Sprior} into Eq.~\eqref{eq:def:Spost} and taking $\eta^2 / \sigma^2 \ll 1$, we obtain an update from $\lpost(t)$ to $\lpost(t+1)$,
\begin{align}
  \lpost(t+1) &= \frac{1}{\frac{1-\eta^2/\sigma^2}{\lpost(t)} + \eta^2} + g^2(t)
  \intertext{assuming $\lpost$ has reached fixed-point, we have $\lpost = \lpost(t) = \lpost(t+1)$,}
  \lpost &= \frac{1}{\frac{1-\eta^2/\sigma^2}{\lpost} + \eta^2} + \ab{g^2}.
  \intertext{Rearranging,}
  \lpost &= \frac{\lpost}{1-\eta^2/\sigma^2 + \eta^2 \lpost} + \ab{g^2}.
  \intertext{Assuming that the magnitude of the update to $\lpost$ is small, we can take a first-order Taylor of the first term,}
  \lpost &\approx \lpost\b{1 + \frac{\eta^2}{\sigma^2} - \eta^2 \lpost} + \ab{g^2}.
  \intertext{cancelling,}
  0 &\approx \eta^2 \lpost^2 - \frac{\eta^2}{\sigma^2} \lpost - \ab{g^2},
  \intertext{and rearranging,}
  0 &\approx \sigma^2 \lpost^2 - \lpost - \frac{\ab{g^2} \sigma^2}{\eta^2},
\end{align}
and finally substituting for $\lpost$ gives the expression in the main text.

\section{Additional data figures}
\begin{figure*}
  \centering
  \includegraphics{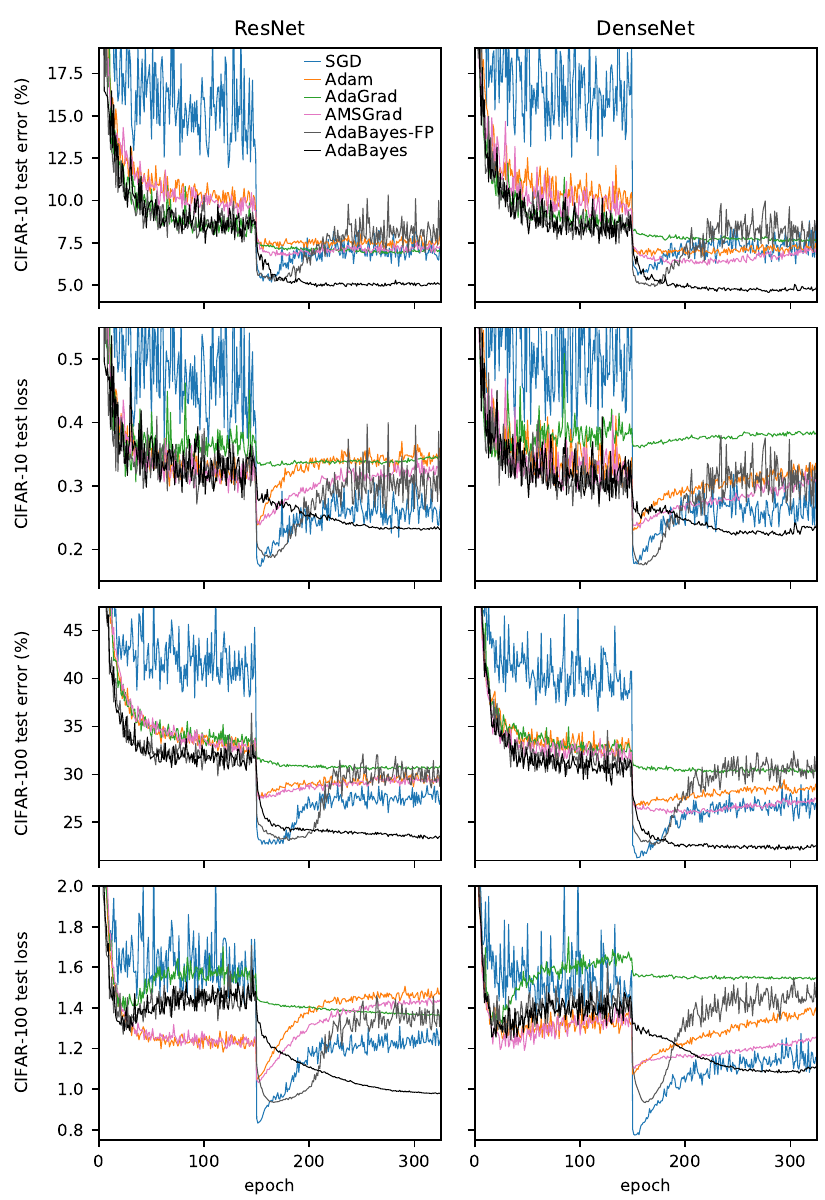}
  \caption{Test loss and classification error for CIFAR-10 and CIFAR-100 for a Resnet-34 and a DenseNet-121, comparing our methods (AdaBayes and AdaBayes-FP) with standard baselines (SGD, \adam{}, AdaGrad and AMSGrad \citep{reddi2018convergence}).
    \label{fig:baselines}
  }
\end{figure*}

\begin{figure*}
  \centering
  \includegraphics{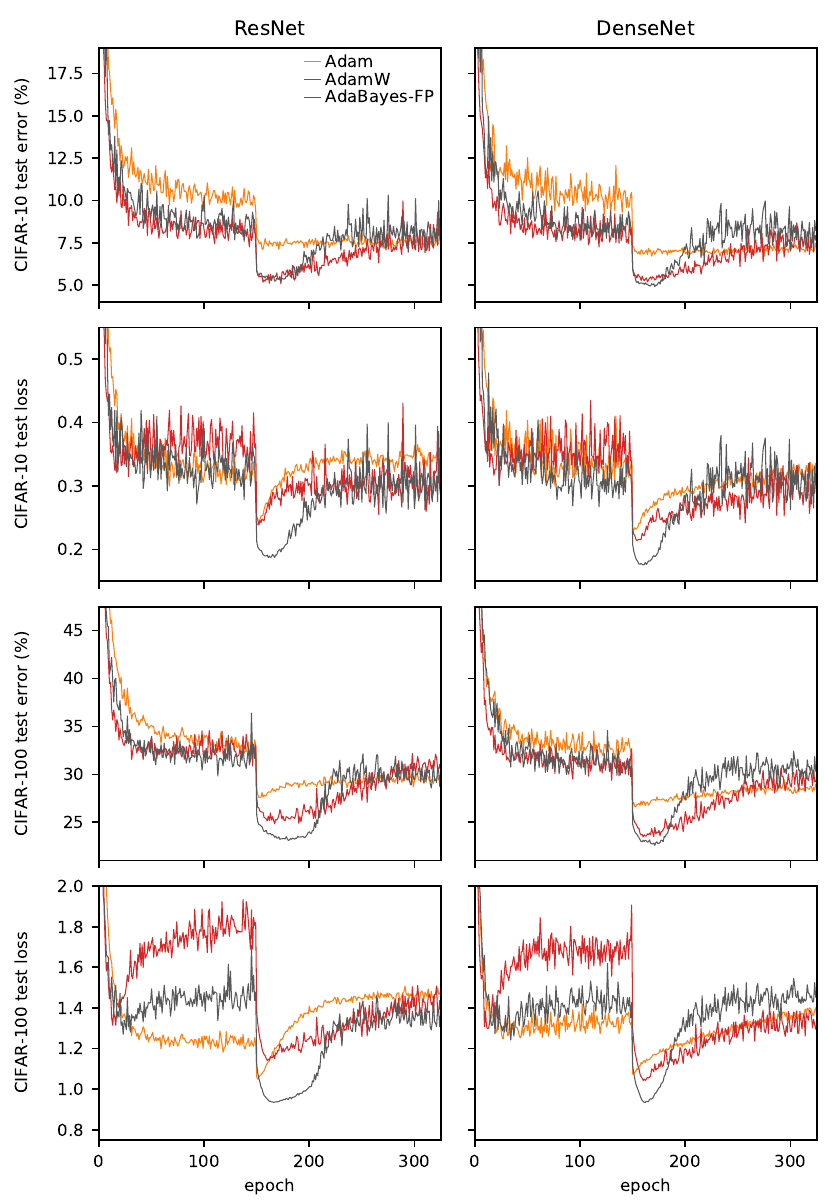}
  \caption{Test loss and classification error for CIFAR-10 and CIFAR-100 for a Resnet-34 and a DenseNet-121, comparing \adam{}, \adamw{} and AdaBayes-FP.
    \label{fig:adamw}
  }
\end{figure*}

\begin{figure*}
  \centering
  \includegraphics{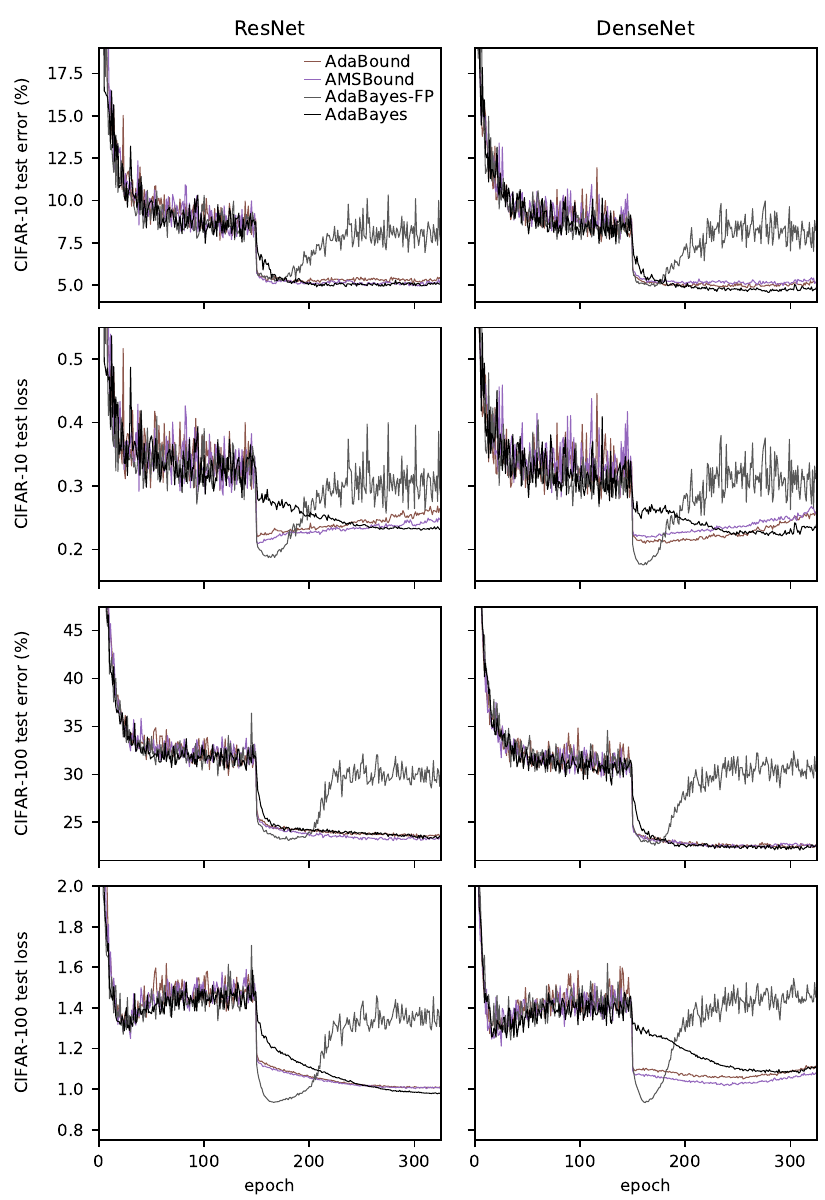}
  \caption{Test loss and classification error for CIFAR-10 and CIFAR-100 for a Resnet-34 and a DenseNet-121, comparing our methods (AdaBayes and AdaBayes-FP) with AdaBound/AMSBound \cite{luo2019adaptive}.
    \label{fig:bound}
  }
\end{figure*}

\begin{figure*}
  \centering
  \includegraphics{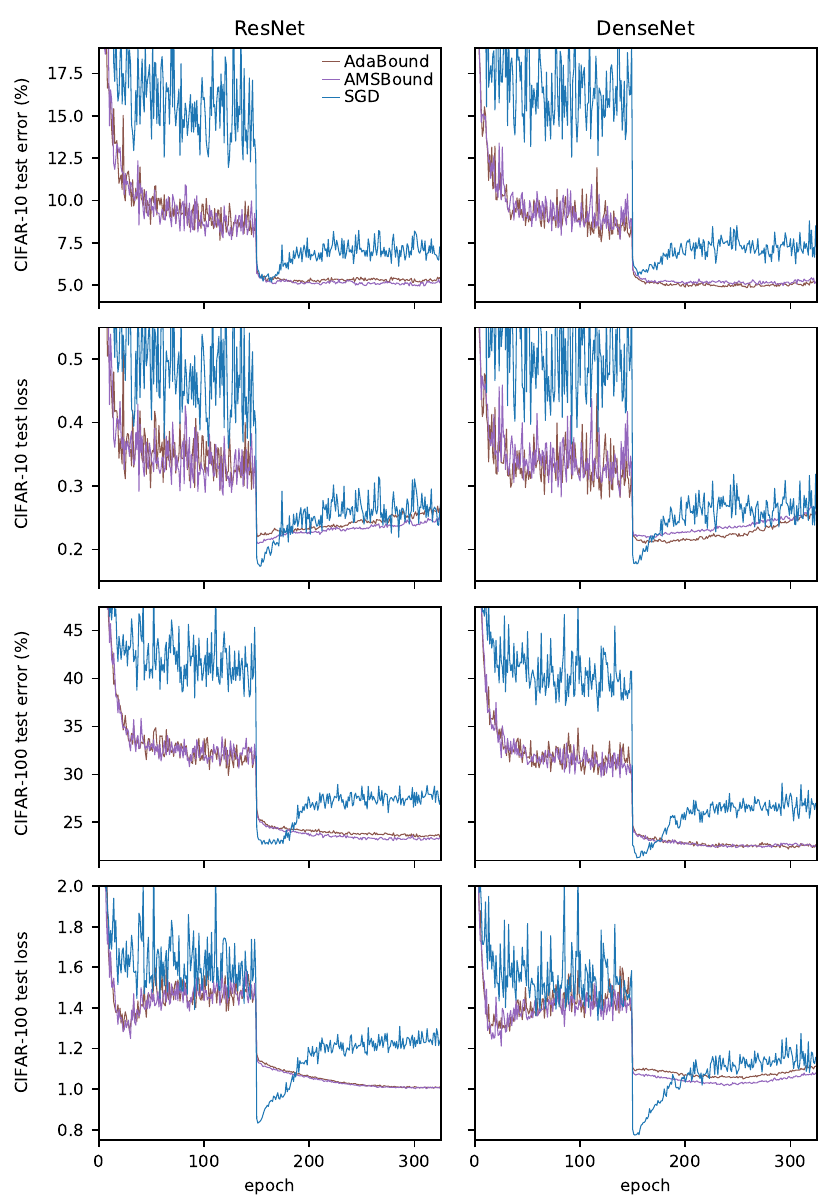}
  \caption{Test loss and classification error for CIFAR-10 and CIFAR-100 for a Resnet-34 and a DenseNet-121, comparing AdaBound/AMSBound \cite{luo2019adaptive} and SGD.
    \label{fig:bound_sgd}
  }
\end{figure*}

\begin{figure*}
  \centering
  \includegraphics{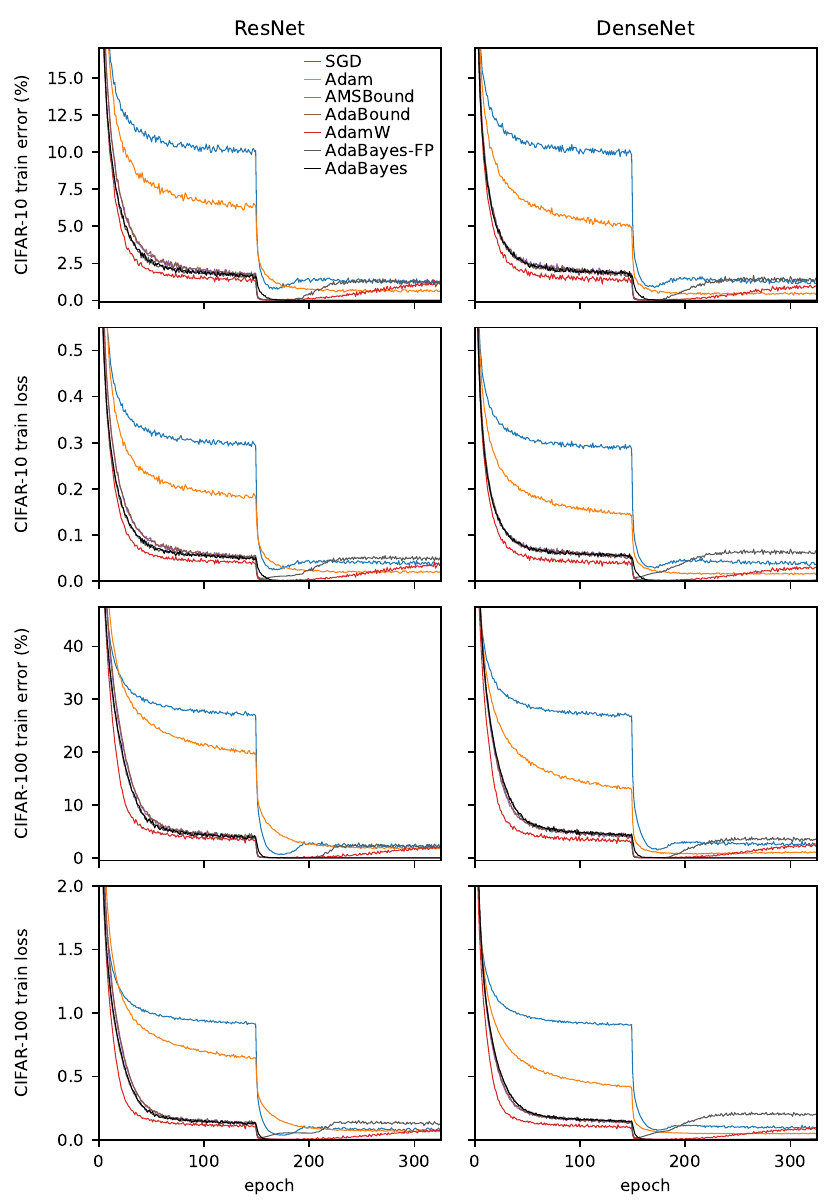}
  \caption{Train loss and classification error for CIFAR-10 and CIFAR-100 for a Resnet-34 and a DenseNet-121, for all methods in Fig.~\ref{fig:all}.
    \label{fig:train}
  }
\end{figure*}

Here, we replot Fig.~\ref{fig:all} to clarify particular comparisons.
In particular, we compare AdaBayes(-FP) with standard baselines (Fig.~\ref{fig:baselines}), \adam{} \adamw{} and AdaBayes-FP (Fig.~\ref{fig:adamw}), AdaBayes(-FP) and Ada/AMSBound (Fig.~\ref{fig:bound}), and Ada/AMSBound and SGD (Fig.~\ref{fig:bound_sgd}).
Finally, we plot the training error and loss for all methods (Fig.~\ref{fig:train}; note the loss does not include the regularizer, so it may go up without this being evidence of overfitting).

\clearpage

\end{document}